\documentclass[11pt]{article}
\usepackage{enumitem}
\usepackage[preprint]{acl}

\usepackage{times}
\usepackage{latexsym}
\usepackage{colortbl}
\usepackage[T1]{fontenc}

\newcommand{\header}[1]{\vspace{1.5mm}\noindent\textbf{#1}.}

\usepackage[utf8]{inputenc}

\usepackage{microtype}

\usepackage{inconsolata}
\usepackage{amsmath, amssymb, amsfonts}
\usepackage{graphicx}

\usepackage{booktabs}

\usepackage{pifont}
\usepackage{subcaption}

\usepackage[most]{tcolorbox}
\newtcolorbox{mybox}[1][]{%
  enhanced,
  breakable,             
  boxrule=0.4pt,
  colframe=blue!35,
  colback=blue!4,        
  arc=3mm,               
  outer arc=3mm,
  left=2.5mm,right=2.5mm,top=2mm,bottom=2mm,
  #1                     
}

\title{EvoIdeator: Evolving Scientific Ideas through Checklist-Grounded Reinforcement Learning}

\author{
    Andreas Sauter\textsuperscript{\rm 1,2} \qquad 
    Yuyue Zhao\textsuperscript{\rm 1} \qquad 
    Jacopo Urbani\textsuperscript{\rm 1,2} \qquad 
    Wenxiang Hu\textsuperscript{\rm 1} \\[0.5ex] 
    {\bf Zaiqiao Meng\textsuperscript{\rm 1} \quad 
    Lun Zhou\textsuperscript{\rm 1} \quad 
    Xiaohui Yan\textsuperscript{\rm 1}\thanks{Corresponding authors.} \quad 
    Yougang Lyu\textsuperscript{\rm 1}\footnotemark[1]} \\[1ex]
    \textsuperscript{\rm 1}Huawei Technologies Co., Ltd. \qquad 
    \textsuperscript{\rm 2}Vrije Universiteit Amsterdam
}

\begin{document}
\maketitle

\begin{abstract}

Scientific idea generation is a cornerstone of autonomous knowledge discovery, yet the iterative evolution required to transform initial concepts into high-quality research proposals remains a formidable challenge for Large Language Models (LLMs). Existing Reinforcement Learning (RL) paradigms often rely on rubric-based scalar rewards that provide global quality scores but lack actionable granularity. Conversely, language-based refinement methods are typically confined to inference-time prompting, targeting models that are not explicitly optimized to internalize such critiques. To bridge this gap, we propose \textbf{EvoIdeator}, a framework that facilitates the evolution of scientific ideas by aligning the RL training objective with \textbf{checklist-grounded feedback}. EvoIdeator leverages a structured judge model to generate two synergistic signals: (1) \emph{lexicographic rewards} for multi-dimensional optimization, and (2) \emph{fine-grained language feedback} that offers span-level critiques regarding grounding, feasibility, and methodological rigor. By integrating these signals into the RL loop, we condition the policy to systematically utilize precise feedback during both optimization and inference. Extensive experiments demonstrate that EvoIdeator, built on Qwen3-4B, significantly outperforms much larger frontier models across key scientific metrics. Crucially, the learned policy exhibits strong generalization to diverse external feedback sources without further fine-tuning, offering a scalable and rigorous path toward self-refining autonomous ideation.
\end{abstract}

\section{Introduction}

The automated generation of novel, high-impact research ideas stands as a frontier challenge in the pursuit of autonomous scientific discovery. Because scientific quality is multifaceted and lacks a single ground truth, reinforcement learning (RL) has emerged as a natural fit for this automated idea generation task, allowing models to optimize complex objectives through qualitative reward signals beyond simple imitation~\cite{bai2022constitutional, ouyang2022training}. However, existing approaches suffer from a fundamental dual gap. 

RL-based methods for scientfic idea generation typically rely on rubric-based scalar rewards that can quantify idea quality but do not specify \emph{which} aspects to change or \emph{how} to improve a given proposal.  These methods optimize long-horizon research behavior through scalar reward signals~\cite{jin2025searchr1, guo2025deepseek} by internalizing broad quality patterns and amortizing the cost of iterative search, but remain confined to a scalar score that omits fine-grained feedback.
Conversely, language feedback methods, which supply fine-grained, actionable critiques in feedback cycles rather than updating the model's weights, are largely restricted to inference-time prompting and target models that are not explicitly trained to leverage such signals~\cite{yamada2025ai, baek2025researchagent, wang2024scimon, su2024two}. 

This dichotomy of approaches undermines alignment between training and inference-time objectives that have been shown to aid performance~\cite{balashankar2025infalign}. Importantly, no existing approach jointly trains a model on scalar RL rewards while receiving fine-grained language feedback in a principled way. This raises the question of whether we can improve scientific ideation by leveraging LLMs' intrinsic capability to follow feedback~\cite{madaan2023self} by explicitly aligning it with our training objectives.

\begin{figure*}[t]
    \centering
    \includegraphics[width=\textwidth]{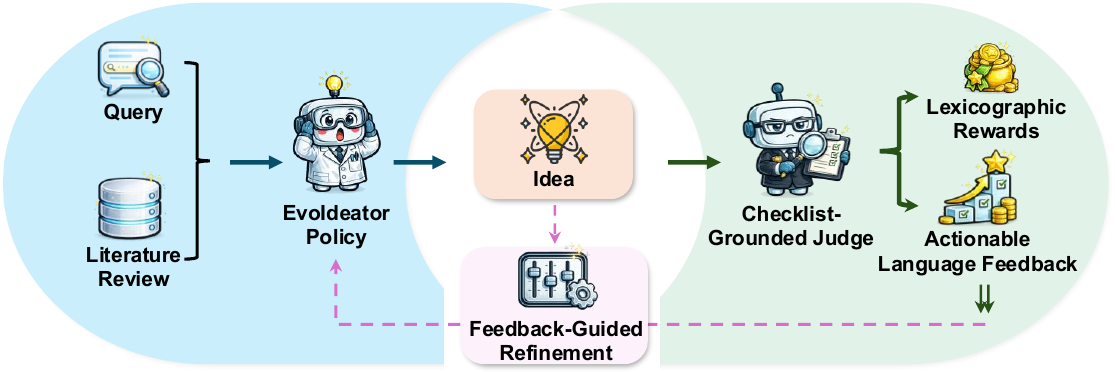}
    \caption{
    Overview of EvoIdeator. Given a research query and a relevant literature review, the EvoIdeator policy generates an initial candidate idea. The idea is then evaluated by a checklist-grounded judge, which produces two complementary signals: lexicographic rewards for multi-dimensional optimization and actionable language feedback that specifies how the idea should be revised. The language feedback, together with the current idea, is fed into the policy revision module to produce an improved idea. In parallel, the lexicographic rewards are used in an RL loop to update the EvoIdeator policy, aligning train-time optimization with inference-time refinement.
    }
    \vspace*{-1mm}
    \label{fig:overview}
    \vspace*{-5mm}

\end{figure*}

To close the dual gap of these  approaches for scientific idea generation, we introduce EvoIdeator, a framework that explicitly aligns train-time RL with checklist-grounded feedback. Building on the Dr.\ GRPO estimator~\cite{liu2025understandingr1zeroliketrainingcritical}, EvoIdeator couples training with an inference-time idea-review cycle in which a judge model delivers two complementary signals: (1) \emph{lexicographic scalar rewards} and (2) \emph{actionable language feedback} that identifies specific failures and provides span-level critiques specifying \emph{which} passages to revise and \emph{how}~\cite{pryzant2023automatic, yuksekgonul2025optimizing}, derived from a structured checklist spanning grounding, feasibility, and methodological rigor. By integrating both signals directly into the training loop, we condition the policy to interpret and execute precise feedback during both optimization and inference. 

In our experiments, we show that coupling the two loops around the same objective and feedback channel indeed yields \emph{additive} quality gains in scientific ideation. Specifically, we show that this leads to EvoIdeator outperforming significantly larger frontier models on key scientific criteria. In summary, our main contributions are:
\begin{itemize}[leftmargin=*,nosep]
    \item We propose EvoIdeator, a framework that aligns train-time RL with language feedback for scientific idea generation, 
    closing the misalignment between training objectives and inference-time evolution.
    
    \item We introduce a dual-signal, checklist-grounded training mechanism combining \emph{lexicographic scalar rewards} 
    with \emph{actionable language feedback} that provides span-level critiques. This coupling yields additive quality gains beyond either signal alone.
    
    \item We demonstrate that EvoIdeator, trained on Qwen3-4B, outperforms significantly larger frontier models on key scientific criteria. We further show that the trained policy can generalize to diverse feedback sources without additional fine-tuning, validating a plug-and-play path toward self-refining ideation.
\end{itemize}

\section{Preliminaries}

\header{Scientific Idea Generation}
Scientific idea generation aims to produce novel research proposals given a research context. Let $\pi_\theta$ denote the actor LLM parameterized by $\theta$. Given an input context $p$ (a research query paired with a literature review), the policy auto-regressively generates a proposal $y=(y_1,\dots,y_T)$:
\begin{align}
\pi_\theta(y \mid p) = \prod_{t=1}^{T} \pi_\theta(y_t \mid p, y_{<t}).
\end{align}
Unlike tasks with verifiable ground-truth answers, scientific ideation requires simultaneously satisfying multiple qualitative desiderata such as grounding, feasibility, and methodological rigor.

\header{Inference-Time Refinement}
A common strategy for improving proposals is iterative refinement conditioned on external feedback. Given an initial proposal $y^{(0)} \sim \pi_\theta(\cdot \mid p_0)$, a feedback mechanism produces critiques $f^{(0)}$. At each step $k$, a composite prompt $p_{k}$ aggregates the original context, the previous draft, and the feedback:
\begin{align}
y^{(k)} \sim \pi_\theta(\cdot \mid p_k), \quad p_k = [p_0;\, y^{(k-1)};\, f^{(k-1)}].
\end{align}
Existing approaches typically apply this loop only at inference time, targeting models not optimized to internalize such feedback.

\section{Method}
\label{sec:method}
In this section, we detail the EvoIdeator method in Figure~\ref{fig:overview}.
First, we introduce the checklist-grounded judge.
Then, we introduce the lexicographic reward scheme.
Next, we introduce the actionable language feedback mechanism.
Finally, the training procedure is explained.

\subsection{Checklist-Grounded Judge}\label{sec:judge}
To provide structured evaluation signals without prohibitive human annotation costs, we employ a single LLM as a judge~\citep{bai2022constitutional}, providing both lexicographic rewards and language feedback. We design a 9-item checklist to ensure fine-grained signals targeting scientific requirements, similar to \cite{goel2025trainingaicoscientistsusing}.

\begin{mybox}[title=Scientific Idea Checklist]
\ttfamily\small
    \textbf{Layout: } Starts with **Title** and uses exactly the required sections in order: Title, Core Problem, Approach, Experimental Plan / Evaluation, Expected Outcomes and Impact, Limitations and Risks, Related Work and Gap.\\
      \textbf{Grounding: } Grounds the idea in established principles / mechanisms / closest SOTA approaches, then articulates their limiting assumption or missing element that creates a real unsolved gap.\\
       \textbf{Feasibility: } Execution is practically feasible: required data / compute / equipment / access are realistic, key dependencies are stated, and the plan does not rely on 'heroic' assumptions.\\
    \textbf{Problem: } States a specific, scoped research question or falsifiable hypothesis and explains why it matters and what changes if it succeeds (scientific / technical / societal impact).\\
     \textbf{Risk: } Identifies key assumptions and failure modes and provides mitigation strategies or alternative paths (Plan B/off-ramps) if core components fail.\\
      \textbf{Method: } Proposes a method that can actually test/falsify the claim and specifies a concrete evaluation plan: datasets / benchmarks / setup, baselines / controls, metrics, and at least one ablation / sanity check, with a validation / statistics / robustness plan where applicable.\\
          \textbf{Writing: } Reads as a professional, self-contained academic abstract/proposal with high specificity: uses concrete named methods / architectures / datasets / formalisms and avoids vague placeholders and fluff.
     \textbf{Innovation: } Clearly states what is new (mechanism / theory / architecture / measurement / data / protocol), is non-obvious (surprising or assumption-challenging), and is expected to generalize beyond a single benchmark / case.\\
          \textbf{Length: }    Respects length constraints: Title = 1 line; Core Problem / Outcomes / Related Work = 1–3 sentences each; Approach = 3 sentences; Experimental Plan / Evaluation = roughly 2 sentences; Limitations = exactly 3 bullet points.
\end{mybox}
For each generated idea, the judge evaluates each checklist item independently to prevent the evaluation of one item from biasing another (e.g., high writing quality masking poor feasibility). For each criterion $c_k$ ($k=1,\dots,m$), the judge generates a binary score $sr_k\in\{0,1\}$ indicating whether the criterion is satisfied in proposal $y$, forming the score vector $\mathbf{sr}(y)=(sr_1,\dots,sr_m)$. For every criterion that is not met ($sr_k=0$), the judge additionally produces a feedback directive $(s_k, l_k, \delta_k)$, where $s_k$ is the offending span, $l_k$ is a description of the issue, and $\delta_k$ is a corrective edit. Note that while the judge model produces reasoning traces, we discard them and only process the final feedback string. The relevant prompts can be found in Appendix \ref{app:prompts_judge}.

By decomposing the concept of scientific quality into discrete verification checkpoints, the judge acts as a translator between abstract desiderata and the reward signals required for policy optimization. This design exploits the fundamental asymmetry between generation and verification: verifying if a specific  constraint (e.g., "is the ablation plan concrete?") is met is computationally significantly more tractable than generating a novel solution from scratch \cite{zeng2025pushingtesttimescalinglimits,goel2025trainingaicoscientistsusing}.

\subsection{Lexicographic Rewards}\label{sec:lexi}
To translate the binary score vector into a scalar training signal that respects the priority structure of scientific evaluation, we employ a lexicographic reward scheme. Generating scientific ideas requires balancing multiple, sometimes competing desiderata (e.g., novelty vs.\ feasibility). Multi-Objective RL (MORL) models the reward as a vector~\citep{roijers2013morl}, but linear scalarization struggles as the number of objectives increases~\citep{ishibuchi2018manyobjectivve}. Lexicographic RL bypasses this by imposing a strict hierarchy: primary objectives must be satisfied before secondary ones are considered~\citep{Gabor1998morl}.

Specifically, we designate $c_1, \dots, c_n$ from the checklist as primary objectives (Grounding, Feasibility, Problem, Risk, Method; $n{=}5$) and $c_{n+1}, \dots, c_m$ as secondary objectives ($m{=}9$). We define the reward $r(\mathbf{sr})$ as:
\begin{equation}
    r(\mathbf{sr}) =   \begin{cases}
        \sum_{i=1}^n sr_i, & \textit{if } \sum_{i=1}^n sr_i \geq n - 1\\
        \sum_{i=1}^m sr_i, & \textit{otherwise}
    \end{cases}
\end{equation}
When nearly all primary objectives are met ($\geq n{-}1$), the reward focuses exclusively on the primary sum, preventing secondary objectives from diluting the signal. Otherwise, the full sum over all $m$ items provides a denser reward that encourages progress on secondary objectives during early training.

\subsection{Actionable Language Feedback}\label{sec:feedback}
To complement scalar rewards with fine-grained guidance for iterative refinement, we introduce actionable language feedback. Building on the \emph{textual gradients} paradigm~\citep{pryzant2023automatic,yuksekgonul2025optimizing}, EvoIdeator generates \emph{checklist-grounded language feedback}: structured critiques derived from the evaluation checklist (Section~\ref{sec:judge}). Each feedback directive localizes a specific issue and specifies a minimal corrective edit. Formally, let $\mathcal{F}(y)=\{k : sr_k=0\}$ be the set of unsatisfied criteria for proposal $y$. The aggregated feedback is the collection of all corresponding directives:
    \begin{align}
    f(y)=\{(s_k, l_k, \delta_k)\}_{k \in \mathcal{F}(y)},
    \end{align}
where $s_k$, $l_k$, and $\delta_k$ are the offending span, issue description, and corrective edit produced by the judge for criterion $c_k$ (Section~\ref{sec:judge}). Unlike generic textual gradients, our feedback is anchored to an explicit checklist that spans grounding, feasibility, and methodological rigor, ensuring that each directive targets a scientifically meaningful criterion.

\subsection{Training}\label{sec:training}
To integrate both signals into a unified RL loop aligned with the inference-time draft--judge--revise procedure, we adopt the Dr.\ GRPO estimator~\cite{liu2025understandingr1zeroliketrainingcritical}.

\header{Multi-Step Rollout}
For each input context $p_0$, we execute a $K$-step rollout following the iterative refinement formulation in the Preliminaries. At step $k{=}0$, the policy generates an initial proposal $y^{(0)} \sim \pi_\theta(\cdot \mid p_0)$. The judge evaluates $y^{(0)}$ to produce the score vector $\mathbf{sr}(y^{(0)})$ and language feedback $f(y^{(0)})$. At each subsequent step $k \geq 1$, the policy generates a revised proposal $y^{(k)} \sim \pi_\theta(\cdot \mid p_k)$ with $p_k = [p_0;\, y^{(k-1)};\, f(y^{(k-1)})]$, and the judge re-evaluates the updated proposal. The cumulative return for a rollout is:
\begin{equation}
R = \sum_{k=0}^{K-1} r\!\left(\mathbf{sr}(y^{(k)})\right),
\end{equation}
where $r(\cdot)$ is the lexicographic reward defined in Section~\ref{sec:lexi}.

\header{Training Objective}
For each prompt $p_0$, we sample $G$ independent rollouts, each yielding a cumulative return $R_j$. Following Dr.\ GRPO, we compute the advantage without length normalization or within-group standardization to avoid verbosity bias:
\begin{equation}
A_j = R_j - \frac{1}{G}\sum_{i=1}^{G} R_i.
\end{equation}
The policy is updated via the clipped surrogate objective with a KL penalty to the reference policy $\pi_{\mathrm{ref}}$:
\begin{align}
\mathcal{L}(\theta) = \mathbb{E}\Big[&\min\!\big(\rho_j A_j,\;\mathrm{clip}(\rho_j, 1{-}\epsilon, 1{+}\epsilon)\, A_j\big) \notag\\
&- \beta\,\mathrm{KL}\!\big(\pi_\theta \,\|\, \pi_{\mathrm{ref}}\big)\Big],
\end{align}
where $\rho_j(\theta) = \pi_\theta(y_t \mid p, y_{<t}) / \pi_{\theta_{\mathrm{old}}}(y_t \mid p, y_{<t})$ is the per-token importance ratio. The advantage $A_j$ is assigned to every token across all $K$ steps of rollout $j$, including intermediate reasoning tokens.

\section{Experiments}\label{sec:exp_setup}
\subsection{Research Questions}
We aim to answer the following research questions in our experiments:
\textbf{RQ1}: Does EvoIdeator outperform the unaligned base model and state-of-the-art reasoning engines on scientific ideation quality?
\textbf{RQ2}: Do EvoIdeator's train-time RL and inference-time language feedback combine additively?
\textbf{RQ3}: Does EvoIdeator's learned feedback protocol generalize to different judge models?
\subsection{Dataset Construction}\label{sec:dataset_construction}
To initialize our RL process ($p_0$), we construct a  dataset of paired \emph{(query, literature\_review)} examples. We rely on a custom pipeline because existing resources are structurally incompatible with train-time RL, being designed primarily as post-hoc evaluation benchmarks rather than large-scale generative training corpora~\cite{moussa2025scholareval,shahid2025literature,qiu2025ai,gu2024generation} or they focus on providing execution environments or metric definitions rather than the pre-computed, retrieval-grounded contexts necessary for training~\cite{zhang2025innogymbenchmarkinginnovationpotential,guo25IdeaBench,he2025shapes}.
Consequently, we implement the following pipeline to generate a scalable dataset of valid training seeds. Find the relevant prompts in Appendix~\ref{app:prompts_dataset}.

\header{Seed Paper Sampling} We first sample a set of recently published seed works from OpenAlex~\cite{priem2022openalex}. We use a fixed random seed to select 1000 accepted works published in 2025 across all domains. To bias towards higher quality, we restrict the selection to journal and conference venues, exclude retracted records, and require an available abstract.

\header{Query and Keyword Generation.} For each seed work, we generate a specific research question using Llama 3~\cite{grattafiori2024llama}\footnote{We use RedHatAI's FP8-dynamic quantization for efficiency: \url{https://huggingface.co/RedHatAI/Llama-3.3-70B-Instruct-FP8-dynamic}}. The model is prompted with the seed paper's title and abstract to generate the query that would have plausibly led to this paper question. Simultaneously, we generate a separate retrieval-oriented keyword string, prompting the model to extract the main keywords from the title and generated question.

\header{Literature Review Synthesis.} We retrieve related literature via Semantic Scholar~\cite{Kinney2023TheSS}. 
We retrieve up to 20 papers per query, discarding questions where fewer than 5 relevant papers are returned. Finally, we synthesize a concise literature review conditioned on the research question and the retrieved abstracts. The LLM is instructed to produce 1-2 compact paragraphs describing common themes and gaps without proposing new solutions.

\subsection{Evaluation Settings}

We split our dataset form Seciton~\ref{sec:dataset_construction} into a training and test set and evaluate all models on 96 \texttt{(query, literature\_review)} from the held-out test set. Each model is assessed on 9 checklist items (Grounding, Feasibility, Problem, Risk, Method, Writing, Innovation, and Length), which are divided into Primary Objectives (the first 5, capturing scientific rigor) and Secondary Objectives (the remaining 4, capturing formatting and novelty). We report mean scores with 95\% confidence intervals, discarding samples where the model fails to produce a valid \texttt{<idea>} block.

To separate initial generation quality from feedback utilization capability, we conduct evaluation at the two inference stages:
\begin{itemize}[leftmargin=*,nosep]
\item \textbf{Generation Step}: The model receives the input \texttt{(query, literature\_review)} pair and produces a research idea in a single forward pass, without external feedback.
\item \textbf{Refinement Step}: The model receives its previous generations along with textual gradients from the judge and produces an improved version.
\end{itemize}
This two-stage protocol allows us to quantify how effectively each model exploits structured, actionable feedback during inference.

\subsection{Baselines}
To evaluate the performance of our trained models, we compare against several external baselines representing different scales and reasoning paradigms.
\begin{itemize}[leftmargin=*,nosep]
\item \textbf{Qwen-4B}: The unaligned base model from which our policies are initialized~\cite{qwen3technicalreport}. This represents the zero-shot performance of a compact reasoning model without task-specific training.
\item \textbf{DeepSeek R1 Distill and DeepSeek-V3.2}: Reasoning-optimized models~\cite{guo2025deepseek, deepseekai2025deepseekv32pushingfrontieropen} trained with process supervision to enhance multi-step reasoning capabilities.
\item \textbf{Gemini 3 Flash}: A large-scale general-purpose model with extended reasoning capabilities, representing the performance of frontier models.
\end{itemize}


\begin{table*}[t]
\centering
\resizebox{\textwidth}{!}{%
\begin{tabular}{lcccccccc}
\toprule
& \multicolumn{5}{c}{\textbf{Primary Objective}} & \multicolumn{3}{c}{\textbf{Secondary Objective}} \\
    \cmidrule(lr){2-6} \cmidrule(lr){7-9} 
\textbf{Method} & \textbf{Grounding} & \textbf{Feasibility} & \textbf{Problem} & \textbf{Risk} & \textbf{Method} & \textbf{Writing} & \textbf{Innovation} & \textbf{Length} \\ 
\midrule
    \multicolumn{9}{c}{\cellcolor{gray!15}\textit{Direct Generation Step}} 
\\
Qwen-4B & $.85 \pm .07$ & $.09 \pm .06$ & $.05 \pm .05$ & $.01 \pm .02$ & $.08 \pm .06$ & $.75 \pm .09$ & $.21 \pm .08$ & $.41 \pm .10$ \\ 
DS R1 Distill & $.53\pm .18$ & $.00 \pm .00$ & $.00 \pm .00$ & $.00 \pm .00$ & $.00\pm .00$ & $.19 \pm .14$ & $.00 \pm .00$ & $.16\pm.13$ \\ 
DS v3.2 & $.80\pm .08$&$.19\pm .08$&$.10 \pm .06$&$.00 \pm .00$&$.12 \pm .07$&$.72 \pm .09$&$.31 \pm .09$&$.40\pm.10$ \\ 
Gemini 3 Flash & $.87 \pm .07$ & $.20 \pm .08$ & $.07 \pm .05$ & $.00 \pm .00$ & $.10 \pm .06$ & $.90 \pm .06$ & $\underline{.55} \pm .10$ & $\textbf{.81} \pm .08$ \\ 
EvoIdeator & $\textbf{.99} \pm .02$ & $.19 \pm .08$ & $.06\pm .05$ & $.03 \pm .04$ & $.18 \pm .08$ & $.96 \pm .04$ & $.32 \pm .10$ & $.37 \pm .10$ \\ 
\midrule
    \multicolumn{9}{c}{\cellcolor{gray!15}\textit{Feedback Refinement Step}} 
\\
Qwen-4B & $\underline{.94} \pm .05$ & $.25 \pm .09$ & $\underline{.91} \pm .06$ & $.19 \pm .08$ & $.39 \pm .10$ & $.92 \pm .06$ & $.41 \pm .10$ & $.33 \pm .10$ \\ 
DS R1 Distill & $.59 \pm .18 $ & $.09 \pm .11$ & .$63 \pm .18$ & $.06 \pm .09$ & $.34 \pm .17$ & $.59 \pm .18$ & $.16 \pm .13$ & $.03 \pm .06$ \\ 
DS v3.2 &$.91 \pm .06$&$\textbf{.54} \pm.10$&$.90 \pm .06$&$\underline{.30} \pm .09$&$\textbf{.72} \pm .09$&$.88 \pm.07$&$.42 \pm .10$&.$25 \pm .09$ \\ 
Gemini 3 Flash& $.91 \pm .06$ & $\underline{.53} \pm .10$ & $.90 \pm .06$ & $.16 \pm .07$ & $.48 \pm .10$ & $\underline{.97} \pm .04$ & $\textbf{.60} \pm .10$ & $\underline{.50} \pm .10$ \\ 
EvoIdeator & $\textbf{.99} \pm .02$ & $.31\pm .09$ & $\textbf{.94} \pm .05$ & $\textbf{.35} \pm .10$ & $\underline{.58} \pm .10$ & $\textbf{.99} \pm .02$ & $.47 \pm .10$ & $.18 \pm .08$ \\ 
\bottomrule
\end{tabular}
}
\vspace{-1mm}
\caption{Mean performance scores for generated ideas (n=96, with 95\% CI). We exclude the "Layout" item as all models achieved near-perfect scores on this aspect. The table visually separates Primary Objectives (critical scientific rigor) from Secondary Objectives (formatting and innovation). \textbf{Bold} indicates the best score per column; \underline{underlined} indicates second-best.}
\label{tab:performance}
\vspace{-5mm}
\end{table*}


\subsection{Implementation Details}
Our model is trained for 100 optimization steps with a global batch size of 5 queries per step. For the Dr.\ GRPO advantage estimation, we sample $G=8$ rollouts per query, resulting in 40 trajectories per update step. We use the AdamW optimizer with a learning rate of $1\times 10^{-6}$ and a KL-divergence coefficient $\beta=0.01$ to maintain stability against the reference policy. 

\section{Experimental Results and Analysis}
\label{sec:results_comparison}

\subsection{Main Results (RQ1)}
In this section, we evaluate the quality of our generated ideas, benchmarking our EvoIdeator against the unaligned base model and several state-of-the-art (SOTA) reasoning engines. To ensure our results are robust and not merely artifacts of overfitting to the train-time judge, we employ DeepSeek-V3.2~\cite{deepseekai2025deepseekv32pushingfrontieropen} as the scoring judge while still providing the language feedback with the distilled DeepSeek R1 model.
 Our results are presented in Table~\ref{tab:performance}.\footnote{We caution that DeepSeek-V3.2's high scores could reflect a known self-preference bias~\cite{zheng2023judging,panickssery2024llm}, as the judge is the same model as the generator. We note that the smaller distilled R1 model appears unsuited for this specific zero-shot generation task, consistently underperforming even the unaligned base model.}  Based on the results, we have four main observations:
 
\begin{itemize}[leftmargin=*,nosep]
    \item  \textbf{EvoIdeator's dual-signal training mechanism internalizes scientific criteria.} EvoIdeator employs both lexicographic scalar rewards and actionable language feedback during training. This dual-signal mechanism enables EvoIdeator to consistently outperform the evidently strong, but unaligned Base model (Qwen-4B) across virtually all checklist items in both generation and refinement steps. Notably, in the critical Refinement Step, EvoIdeator achieves a near-perfect Grounding score (.99) and leads in Problem definition (.94 vs .91) and Risk assessment (.35 vs .19), demonstrating that the training pipeline has successfully internalized the scientific criteria.

    \item  \textbf{EvoIdeator's train-inference alignment bridges the capability gap with larger models.} By aligning train-time RL with inference-time feedback loops, EvoIdeator  effectively integrates and executes precise feedback during both optimization and inference. EvoIdeator outperforms Gemini 3 Flash on 4 out of 5 primary objectives (Grounding, Problem, Risk, Method) and surpasses DeepSeek-V3.2 on 3 out of 5 primary objectives after refinement, successfully lifting the base performance clearly above significantly larger models.  This demonstrates that, by closing the misalignment gap, EvoIdeator enables small models to reach state-of-the-art performance.

    \item \textbf{EvoIdeator's lexicographic reward prioritization produces expected trade-offs.} EvoIdeator's lexicographic reward scheme explicitly prioritizes primary objectives (Grounding, Feasibility, Problem, Risk, Method) over secondary objectives (Writing, Innovation, Length). As expected, EvoIdeator excels at scientific rigor while underperforming on Innovation and Length compliance. This confirms that EvoIdeator is behaving exactly as optimized: sacrificing formatting strictness to ensure the scientific idea is sound and falsifiable.

    \item  \textbf{Checklist-grounded feedback benefits all models, but alignment amplifies the gains.} We observe that all models show score increases between the Generation and Refinement steps, confirming the general utility of structured language feedback. However, EvoIdeator, which is explicitly trained to internalize such feedback, achieves the highest post-refinement scores on primary objectives. This supports our core hypothesis that aligning the training objective with the inference-time feedback loop yields gains beyond what either paradigm achieves alone.
\end{itemize}

\subsection{Additive Effects of RL Training and Inference-Time Refinement (RQ2)}
\label{sec:additive_gains}
In this section, we isolate the contributions of train-time weight updates and inference-time language feedback, investigating whether their combination yields the hypothesized additive benefits. To do so, we analyze the performance trajectories of four distinct configurations: The Informed model, as described in Section~\ref{sec:method}; the Non-Informed model (EvoIdeator but without language feedback during training and inference); the Base model (Qwen-4B) with language feedback during inference; and the Non-Informed Base (Qwen-4B) model without language feedback during inference. Figure~\ref{fig:synergies} illustrates the improvement trajectory of the average summed scores ($\sum_{i=0}^m sr_i$) from the initial generation (Step 0) to the refined output (Step 1). Based on the results in Figure~\ref{fig:synergies}, we have three main observations:

\begin{figure}[t]
    \centering
    \includegraphics[width=\columnwidth]{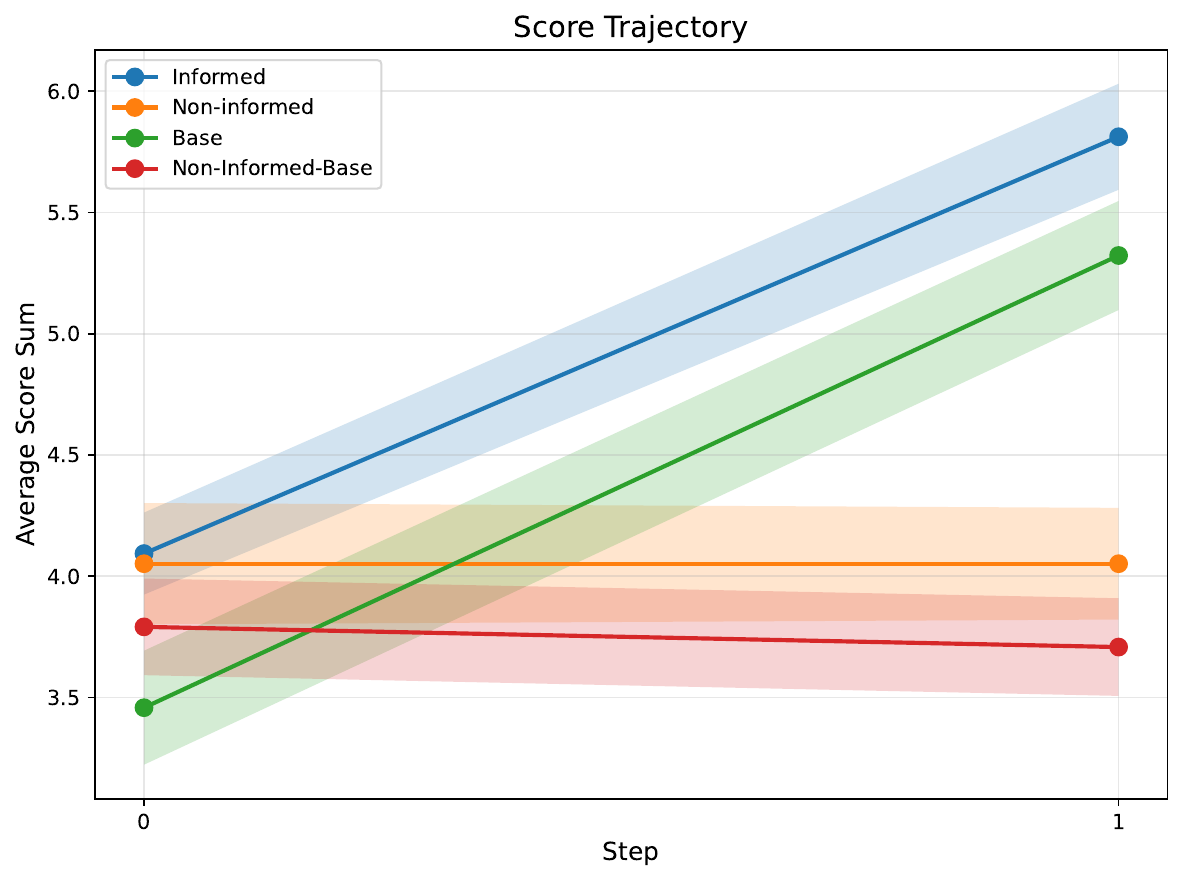}
    \caption{Performance trajectories across two generation steps.  We compare the \emph{Informed} model (Green) against the \emph{Uninformed} model (Orange) and their respective untrained checkpoints (Blue/Gray). \textbf{Step 0} represents the initial zero-shot generation; \textbf{Step 1} represents the output after one round of refinement (via language feedback for Informed/Base, or self-correction for Non-informed). The lines represent the average over the summed scores. Shaded regions denote 95\% confidence intervals.}
    \vspace{-1mm}
    \label{fig:synergies}
    \vspace{-5mm}
\end{figure}

\begin{itemize}[leftmargin=*,nosep]
    \item  \textbf{EvoIdeator's RL formulation elevates initial generation quality through weight-level internalization.} Both the Informed and Non-Informed variants of EvoIdeator exhibit nearly identical high-performance scores at Step 0, outperforming their untrained counterparts (Base and Non-Informed Base). This demonstrates that EvoIdeator's lexicographic reward-driven training successfully distills general scientific criteria into the model weights during the RL phase, enabling superior one-shot generation quality regardless of whether the model will later receive language feedback during inference. The consistent intercept improvement across both trained variants confirms that the reward signal effectively internalizes the checklist criteria as a general generation capability.

    \item \textbf{EvoIdeator's language feedback mechanism enables effective refinement beyond self-correction.} At Step 1, we observe a sharp divergence: EvoIdeator's Informed variant (equipped with language feedback) shows significant improvements in overall quality, while the Non-Informed variant (relying on intrinsic self-correction) stagnates. The same pattern holds for untrained models: Base (with feedback) improves, while Non-Informed Base (without feedback) does not. This empirically demonstrates that for scientific ideation, models cannot effectively refine quality purely through introspection. Structured, actionable language feedback is required to drive meaningful refinement, and training without this feedback results in a model that generates strong initial ideas but cannot exploit iterative improvement opportunities.

    \item  \textbf{EvoIdeator's dual-phase design achieves additive quality gains.} The performance gap between EvoIdeator's Informed variant and the untrained Base model remains constant across the refinement step, indicating a robust additive effect. EvoIdeator benefits from the high initial intercept provided by RL training plus the consistent refinement slope provided by language feedback. This result shows that EvoIdeator's train-time alignment does not compromise the policy's plasticity: the Informed policy retains the ability to incorporate feedback just as effectively as the base model, while starting from a much higher baseline. These findings are generally also reflected in the score trajectories for each individual checklist item, which can be found in Appendix~\ref{app:feedback_effects}.
\end{itemize}

\subsection{Cross-Judge Generalization (RQ3)}
\label{sec:judge_dependencies}
We investigate the robustness of our \emph{Informed} policy to the source of language feedback, specifically testing for overfitting to the training judge (DeepSeek R1 Distill 70B). We evaluate language feedback from a spectrum of out-of-distribution providers: DeepSeek R1 Distill 14B, DeepSeek-V3.2, and Gemini 3 Flash. Results are presented in Table~\ref{tab:judges}.
\begin{table}[t]
\centering
\resizebox{\columnwidth}{!}{%
\begin{tabular}{lcc}
\toprule
\textbf{Feedback} & \textbf{Generation} & \textbf{Refinement}  \\
\textbf{Provider} & \textbf{Step} & \textbf{Step}\\
\midrule
DS R1 14B Distill & $ 4.07 \pm .19$& $5.64 \pm .22$\\
DS R1 70B Distill & $4.09 \pm .17$ &$5.81 \pm.22$\\
DS v3.2 & $4.05 \pm .18$&$6.02 \pm .23$\\
Gemini 3 Flash & $3.97 \pm .19$ & $5.13 \pm .25$\\
\bottomrule
\end{tabular}
}
\caption{Average sum over checklist items and 95\% confidence intervals for different models that provide language feedback to our informed model for each step.}
\label{tab:judges}
\vspace{-5mm}
\end{table}
We have two main observations:
\begin{itemize}[leftmargin=*,nosep]
    \item  \textbf{EvoIdeator's feedback mechanism generalizes within lineages.} Within the DeepSeek family, we observe a clear scaling pattern: refinement scores improve monotonically with provider capability ($14B \rightarrow 70B \rightarrow V3.2$). Despite architectural differences across these models that nullify potential self-preference effects, EvoIdeator successfully transfers its learned feedback interpretation capability, benefiting from superior reasoning without a domain shift. We attribute this successful transfer to a shared post-training lineage: these models likely share RLHF distributions, resulting in a consistent feedback dialect (tone, structure, reasoning style) that EvoIdeator's policy can leverage. This demonstrates that EvoIdeator learns generalizable feedback interpretation patterns rather than overfitting to a specific judge model.

    \item \textbf{EvoIdeator's feedback protocol exhibits dialect sensitivity across model families.} Performance drops significantly when using Gemini 3 Flash as the feedback provider, despite Gemini's frontier-scale capabilities. As Gemini originates from a distinct training and alignment lineage, its feedback likely follows a stylistic distribution unseen during EvoIdeator's training phase. This empirically demonstrates that language feedback functions as a learned communication protocol: while EvoIdeator's policy generalizes to better reasoning content, it remains sensitive to the feedback's stylistic dialect.
\end{itemize}

Consequently, EvoIdeator exhibits plug-and-play adaptability within the same alignment lineage, allowing for inference-time upgrades via more capable judges, while cross-family generalization may require enforcing a standardized feedback format.

\section{Related Work}

The application of LLMs to complex reasoning and scientific ideation broadly falls into three paradigms, mirroring the dual gap identified in our introduction.

\header{Train-Time Optimization.}
A growing body of work treats text improvement as a pure train-time intervention. Most deep-research and scientific ideation systems optimize long-horizon behaviors via RL reward signals \cite{li2024ldc, jin2025searchr1, qi2025webrl, yuan2025memsearchertrainingllmsreason, guo2025deepseek, wan2025pokeeresearch, tongyideepresearchteam2025tongyideepresearchtechnicalreport, qiao2025webresearcher,  wengcycleresearcher, chen2025research_both, team2025kimi, goel2025trainingaicoscientistsusing, li2025jointly}, while \emph{Text2Grad} converts textual feedback into span-wise gradient signals \cite{wang2025text2grad}. These methods internalize research heuristics but rely on scalar rewards that lack actionable granularity.

\header{Inference-Time Refinement.}
Many approaches rely entirely on inference-time scaling to dynamically refine outputs without altering model weights, either by modifying token probabilities \cite{khanovargs, huang2024deal, shi2024decoding, chen2024pad, wangself} or through search algorithms \cite{liu2023making, hung2025reward, park2025ensembling, inoue2025wider}. More closely related to our refinement mechanism are paradigms that iteratively critique and refine drafts using language feedback \cite{yao2023tree, xie2023self, xu2024llmrefine, madaan2023self, shinn2023reflexion, chenteaching, goucritic, litest, lee2025feedback}. Within scientific ideation, several systems adopt pure inference-time loops \cite{yang2024large, wang2024scimon, su2024two, yamada2025ai, baek2025researchagent}. While these methods supply rich feedback, they target models not trained to leverage such signals.

\header{Bridging Train and Inference Time.}
An emerging line combines train-time RL with inference-time procedures. \emph{SCoRe}~\cite{kumartraining} and \emph{PAG}~\cite{jiang2025pag} target correctness via intrinsic self-correction, but \emph{SCoRe} requires compute-heavy stage-wise regularization, while \emph{PAG} treats turns as independent updates, neglecting sequential dependencies. \emph{Critique-GRPO}~\cite{zhang2025critique} incorporates external feedback, but its critique remains unstructured and lacks fine-grained alignment with the training objective.

Unlike these methods, which focus on narrow correctness tasks with unstructured feedback, EvoIdeator closes both halves of the dual gap for scientific ideation: it pairs lexicographic scalar rewards with checklist-grounded language feedback, explicitly aligning train-time RL with the inference-time refinement loop.


\section{Conclusion}
We introduce EvoIdeator, a framework that closes the dual gap between scalar RL rewards and inference-time language feedback for scientific idea generation. By pairing lexicographic rewards with checklist-grounded feedback in a unified RL loop, a compact 4B model outperforms larger frontier models on primary scientific criteria, with additive gains from RL and language feedback and cross-judge generalization without retraining.

\section{Limitations}
Our approach exhibits specific limitations that outline clear targets for future research.
First, our lexicographic reward scheme strictly prioritizes scientific rigor, causing secondary objectives like Innovation and Length to be occasionally deprioritized. Exploring adaptive weighting or Pareto-based MORL strategies that dynamically balance primary and secondary objectives is a promising direction.
While our framework naturally generalizes to longer refinement horizons, investigating how performance scales with additional iterations and how to optimally allocate compute across steps remains an open question.
Third, our evaluation relies on LLM-based judges, which is standard practice in scientific ideation benchmarks. Incorporating expert human evaluation at scale would further strengthen the validity of the results, though this is a shared challenge across the field.


\bibliography{custom}

\newpage
\appendix
\section{Prompts}
\label{app:prompts}

\subsection{Dataset Creation}\label{app:prompts_dataset}
This Section contains the prompts that we used in our dataset creation pipeline in Section \ref{sec:dataset_construction}.
\subsubsection{System Prompts}
These are the three system prompts used together with the respective user prompts.
\begin{mybox}[title=System prompts]
\ttfamily\small
(Question generation)\\
You are a data processing engine. Output only the requested text.\\[0.75em]
(S2 keyword query generation)\\
You are a data processing engine. Output only a keyword query string.\\[0.75em]
(Literature review generation)\\
You are a data processing engine. Output only the requested literature review text.
\end{mybox}

\subsubsection{Research Question Generation Prompt}
This prompt is used to extract a user question from the given title and abstract of a paper in our dataset generation pipeline.
\begin{mybox}[title=Research question generation]
\ttfamily\small
**TASK**\\
Given this title and abstract, generate EXACTLY ONE concise research question\\
that a researcher might have asked to arrive at this research direction.\\[0.75em]
INPUT\\
Title: \{TITLE\}\\[0.25em]
Abstract: \{ABSTRACT\}\\[0.75em]
CONSTRAINTS\\
- The question should not be formulated as a information retrieveing question\\
  (e.g.: "How does X work?" or "What are methods solving X?"), but rather goal\\
  oriented (e.g.: "How can Z be achieved?" or "How can X and Y be used to do Z?").\\
- Output ONLY the question text.\\
- The question should not be too specific or detailted but match a balanced\\
  abstraction level.\\
- ONE question only.\\
- 1 sentence preferred; at most 2.\\
- End with a question mark.\\
- No quotes, no prefixes, no bullet points.\\
- The research question should not be too very specific.\\[0.75em]
OUTPUT
\end{mybox}

\subsubsection{Semantic Scholar Keyword Query Prompt}
We use this prompt to extract the main keywords related to a specific title and research question. These keywords are used to query the Semantic Scholar API for related works.
\begin{mybox}[title=Semantic Scholar keyword query (retrieval-only)]
\ttfamily\small
TASK\\
Create a Semantic Scholar keyword search query that will retrieve papers about\\
the same topic. Extract those keywords that caputre the content the best.\\[0.75em]
INPUT\\
Paper title: \{TITLE\}\\
Research question: \{QUESTION\}\\[0.75em]
CONSTRAINTS\\
- Output ONLY the query string.\\
- 4 keywords (space-separated).\\
- No punctuation, no quotes, no hyphens.\\
- Avoid generic words (e.g., method, approach, study, paper).\\
- Prefer specific domain terms from the title/question.\\[0.75em]
OUTPUT
\end{mybox}

\subsubsection{Literature Review Generation Prompt}
This prompt is used to create a literature review from a research question and a set of works that are related to this question.
\begin{mybox}[title=Literature review generation]
\ttfamily\small
TASK\\
Write a concise literature review based on the research question and the related\\
papers list. The result should be a related work section that summarized the\\
state of the are researcha dn identifies th emain gaps regarding the question.\\[0.75em]
RESEARCH QUESTION\\
\{QUESTION\}\\[0.75em]
RELATED PAPERS (title + abstract snippets)\\
\{PAPERS\_BLOCK\}\\[0.75em]
CONSTRAINTS\\
- Output 1--2 paragraphs total.\\
- No headings, no bullet points.\\
- Focus on: common approaches, key themes, gaps/limitations, and where the\\
  question fits.\\
- Do not fabricate citations; only refer to what's plausible from the provided papers.\\
- Refer to the papers with proper citations (e.g. Author1 et al., 2024)\\
- Keep it compact (roughly 120--250 words).\\
- Avoid introductiory sentences that restate the question.\\
- do not propose solutions to the question\\[0.75em]
OUTPUT
\end{mybox}

\subsection{Judge Prompts}\label{app:prompts_judge}
\subsubsection{System Prompt}
This prompt is used as a system prompt for evaluation of generated scientific prompts.
\begin{mybox}[title=Judge System Prompt]
\ttfamily\small
\#\#\# Task\\
You are a strict, impartial scientific reviewer. Your task is to evaluate a GENERATED IDEA based on a detailed requirement. Your evaluation must be based on a step-by-step reasoning process.\\

\#\#\# Internal Reasoning Process (Chain-of-Thought)\\
You must engage in a chain-of-thought reasoning process. This process should not be part of the final output.

1.  **Analyze the Generated Idea**: Read the idea carefully to understand its core claims, proposed methods, and expected outcomes.

2.  **Evaluate Against Requirement**: Critically assess the idea. Identify specific strengths and weaknesses regarding the requirement.

3.  **Score Justification**: Formulate a clear justification for the score you will assign. Connect your reasoning directly to the criteria in the requirement. Argue why the score could not be different.

4.  **Synthesize Overall Judgement**: Aggregate your dimensional assessments into a final assessment whether the requirement is met or not.

5.  **Identify Actionable Feedback**: If the requirement is not met, provide a "span feedback" as described below. Pinpoint a specific, short excerpt from the generated idea that can be improved strictly regarding the requirment. Devise a concrete suggestion for revision that would directly address a weakness you identified.\\

\#\# Score rules:

- score MUST be an integer: 0, or 1.

- If the requirement is fully met, score is 1; otherwise 0.

- The score block MUST be last.

- 1 is only awarded if the idea clearly meets the requirement.

- If the idea is much too vague or incomplete to judge, give a 0.

- If you are unsure between two scores, select the lower one.\\

\#\# Span feedback block rules:

- span\_text MUST be an exact, contiguous excerpt copied verbatim from the GENERATED IDEA.

- if the GENERATED IDEA is only one word, consider it as missing idea. The span\_text should then be empty.

- Keep span\_text short (prefer 20–160 characters). 

- Each span must have a clear requirement-linked issue and a concrete, actionable improvement suggestion.\\
                                    
\#\#\# Output Requirements (STRICT)

If the score is 0, you MUST output:

- Exactly 1 span feedback block, then

- Exactly 1 score block.

If the score is 1, you MUST output:

- Only the score block. No other output whatsoever.    \\                          

\#\#\# OUTPUT TEMPLATE span feedback block

span\_text: "\textless verbatim excerpt from GENERATED IDEA\textgreater"

issue: "\textless what is weak, explicitly tied to the requirement\textgreater"

suggestion: "\textless specific revision that would satisfy the requirement\textgreater"\\
                                    
\#\#\# OUTPUT TEMPLATE score block

score: <0|1>     
\end{mybox}

\subsubsection{Query Prompt}
This prompt is used in conjuction with the system prompt to evaluate a generated scientific idea with respect to a specific requirement from the checklist described in Section \ref{sec:judge}.
\begin{mybox}[title=Judge Query Prompt]
\ttfamily\small
\#\#\# Input Data 

[START GENERATED IDEA]

{generated\_idea}

[END GENERATED IDEA]\\
                                   
[START REQUIREMENT]

{requirement}

[END REQUIREMENT]
\end{mybox}

\subsection{Idea Generation Prompts}
These prompts are used to instruct the generation LLM to generate new, and refine previously generated ideas.

\subsubsection{Idea Generation System Prompt}
System prompt at the beginning of each idea generation/refinement rollout.
\begin{mybox}[title=Idea Generation System Prompt]
\ttfamily\small
You are a senior research scientist designing a single, high-quality research idea.\\

Objectives (must follow):

- Propose EXACTLY ONE idea.

- Maximize novelty while staying realistically feasible with current methods/data/compute.

- Align tightly with the user’s QUERY and ground claims in the LITERATURE REVIEW when provided.

- Clearly distinguish what is NEW vs. what is established prior work.

- Follow the required output headings and length constraints exactly.\\

Reasoning:

- Think step-by-step privately to (1) parse the QUERY, (2) extract key themes/gaps from the LITERATURE REVIEW, and (3) choose a novel but feasible direction.\\

Style constraints:

- Be specific (methods, data types, baselines, metrics); avoid vague buzzwords.

- Paraphrase the LITERATURE REVIEW; do not copy text.

- Do not refer to yourself, the prompt, or the process.

- Do not mention the PREVIOUS IDEA or FEEDBACK or internal reasoning or the user in the final output.

- You must output the final idea in an <idea></idea> block after the last </think> token.\\

OUTPUT FORMAT (mandatory; use these headings in this exact order):

<idea>

**Title**

- 1 short line with a specific, descriptive title.\\

**Core Problem**

- 1-2 sentences stating the problem + gap (use the LITERATURE REVIEW when available) and why it matters.\\

**Approach**

- Roughly 3 sentences.

- State the core novelty (new formulation/method/evaluation/theory), the main techniques, and the data/resources you would use.

- Include key assumptions/constraints.\\

**Experimental Plan / Evaluation**

- Roughly 2 sentences.

- Specify datasets or collection strategy, baselines, metrics, and at least one ablation/robustness test.\\

**Expected Outcomes and Impact**

- 1-2 sentences on expected results and how this advances the field vs. existing work.\\

**Limitations and Risks**

- Roughly 3 bullet points.

- Include at least one feasibility risk and one conceptual risk.\\

**Related Work and Gap**

- 2-3 sentences summarizing the most relevant prior work (from the LITERATURE REVIEW and general knowledge if needed) and the unresolved gap your idea targets.

</idea>
\end{mybox}

\subsubsection{Initial Idea Generation Prompt}\label{app:prompts_ideagen}
Prompt to generate the initial idea.
\begin{mybox}[title=Initial Idea Generation Query]
\ttfamily\small
\#Input Data

[START QUERY]

{query}

[END QUERY]\\
                                   
[LITERATURE REVIEW]

\{literature\_review\}

[END LITERATURE REVIEW]\\
                                   
\#Task:

Propose a completely new research idea that satisfies the global objectives and quality criteria from the system message according to the generation rules in the system message.\\

\#Generation rules:

- The idea should be as novel as reasonably possible while still feasible.

- It must be closely aligned with the QUERY.

- When LITERATURE REVIEW is non-empty, ground the idea in that prior work and clearly state what is new.

- When LITERATURE REVIEW is EMPTY, rely on general domain knowledge but keep the idea scientifically plausible.\\

Use the output rules from the system message.

Remember: the full idea must be inside the <idea></idea> box.
\end{mybox}

\nocite{lyumacpo,lyu2024knowtuning,lyu2023feature,lyu2022improving,lyu2023multi,zhang2024towards,lyu2025self,lyu2025deepshop,lyu2024cognitive,lyu2026evoscientist,wang2025cooperative,shi2025deep}

\subsubsection{Idea Refinement Prompt}\label{app:prompts_idearefine}
Prompt for further refinement steps.
\begin{mybox}[title=Idea Refinement Query]
\ttfamily\small
\#Input Data

[START FEEDBACK]

{feedback}

[END FEEDBACK]\\

\#Task:

Your task is to revise and improve the PREVIOUS IDEA based on the FEEDBACK while preserving its core contribution whenever possible.\\

\#Revision rules:

- Treat the PREVIOUS IDEA as a draft. Preserve useful structure, key assumptions, and main objectives unless the FEEDBACK explicitly asks to replace them.

- Make targeted edits that directly address each point of "span\_text"s in FEEDBACK from the previous idea. Look at the specific "issue" for that span and implement the corresponding "suggestion".

- If FEEDBACK is EMPTY, make small, local improvements only (clarity, feasibility, better grounding, sharper novelty); do NOT change the core topic or contribution.

- Do NOT switch to a completely new idea unless the FEEDBACK explicitly requests a new direction.

- The revised idea must still satisfy all global objectives and quality criteria from the system message.

- Always output a full, self-contained research idea in the required format, not a diff or a partial edit.

- Never refer to the FEEDBACK or PREVIOUS IDEA in the final text.\\

Use the output rules from the system message.

Remember: the full idea must be inside the <idea></idea> box.
\end{mybox}

\section{Per Criteria Feedback Effect}\label{app:feedback_effects}
\begin{figure*}[t]
    \centering
    \begin{subfigure}{0.47\textwidth}
        \centering
        \includegraphics[width=\linewidth]{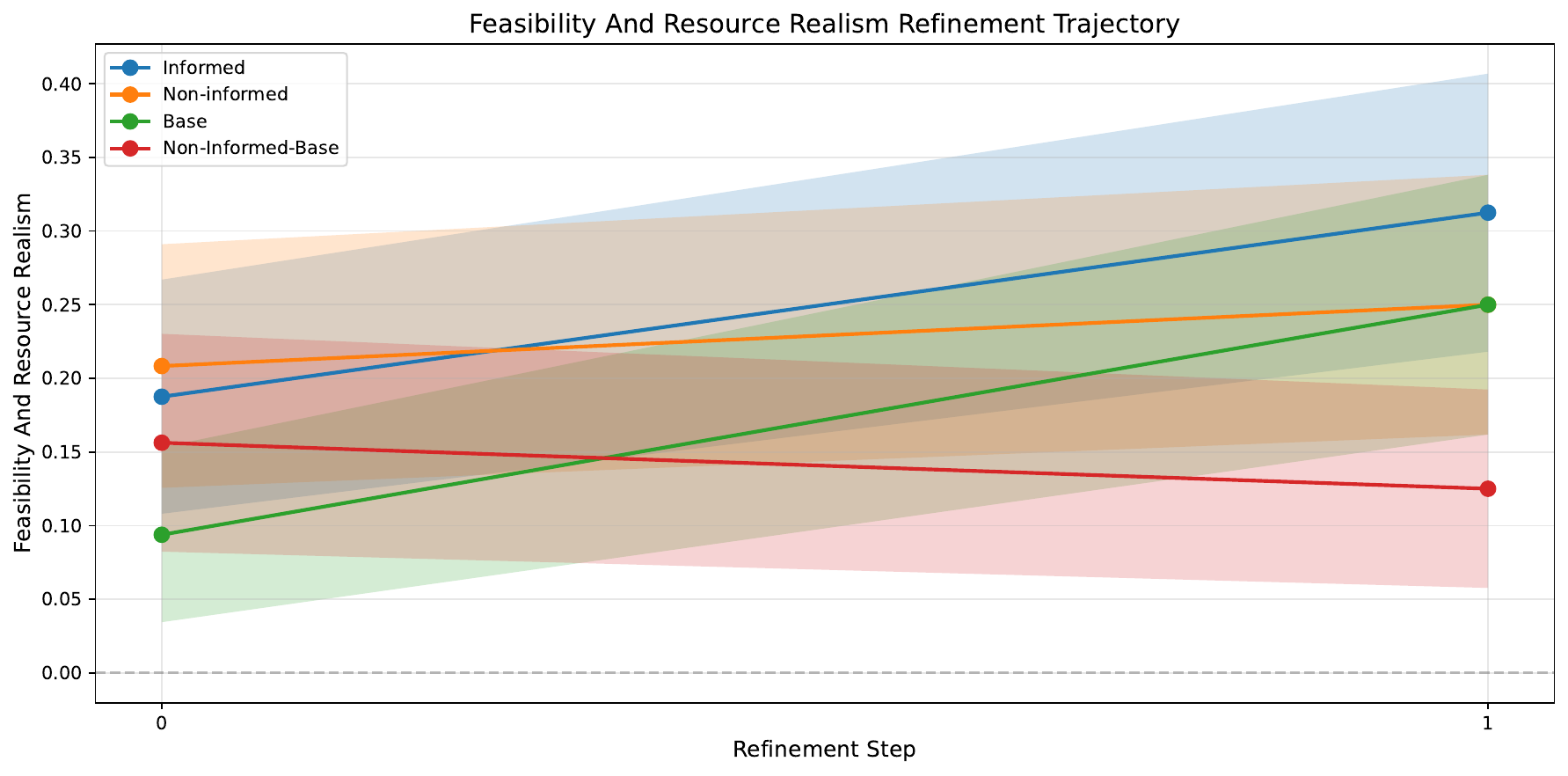}
        \caption{Feasibility and Resource Realism}
    \end{subfigure}%
    \hfill
    \begin{subfigure}{0.47\textwidth}
        \centering
        \includegraphics[width=\linewidth]{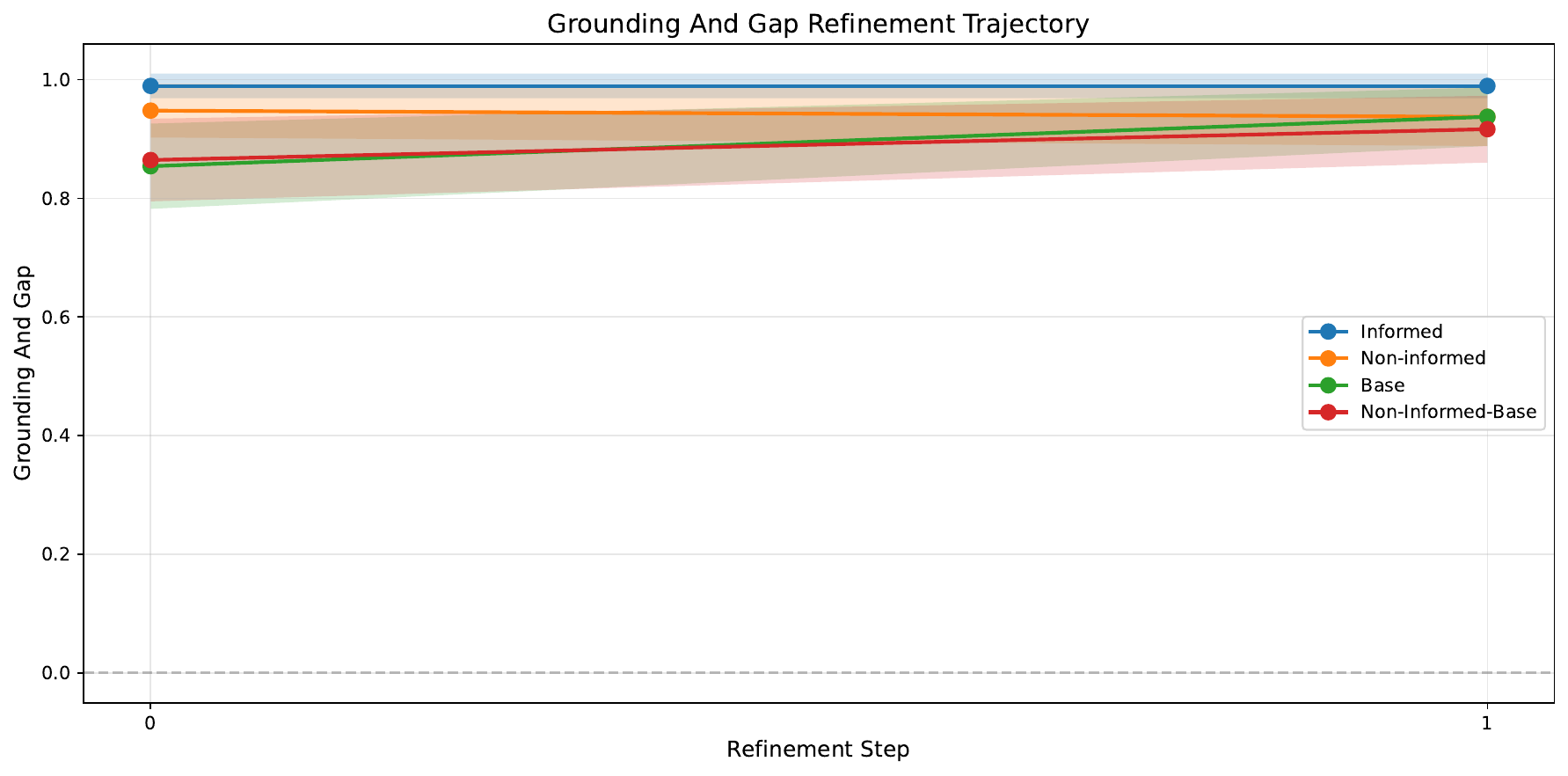}
        \caption{Grounding and Gap}
    \end{subfigure}

    \vspace{1em} 

    \begin{subfigure}{0.47\textwidth}
        \centering
        \includegraphics[width=\linewidth]{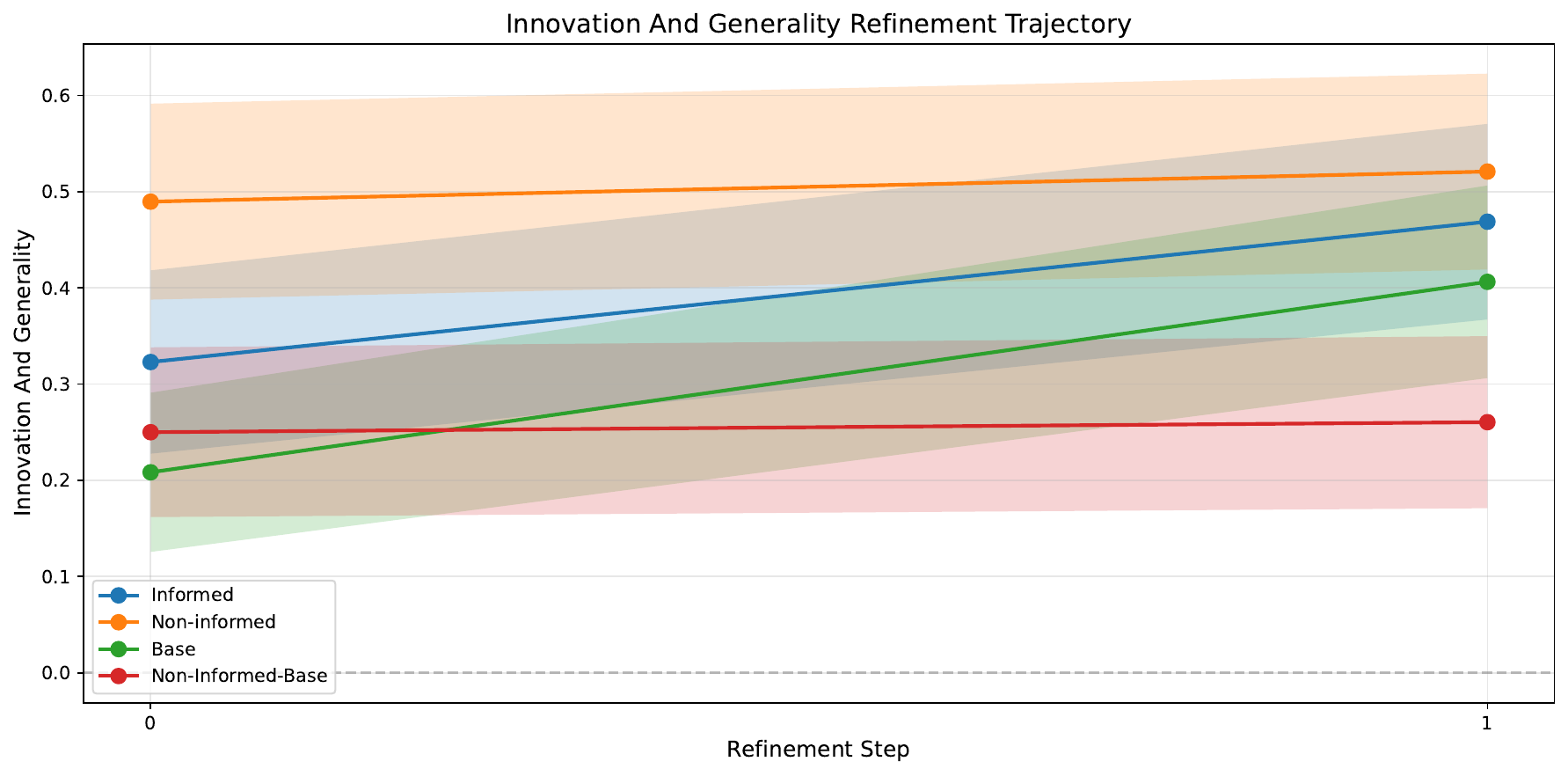}
        \caption{Innovation and Generality}
    \end{subfigure}%
    \hfill
    \begin{subfigure}{0.47\textwidth}
        \centering
        \includegraphics[width=\linewidth]{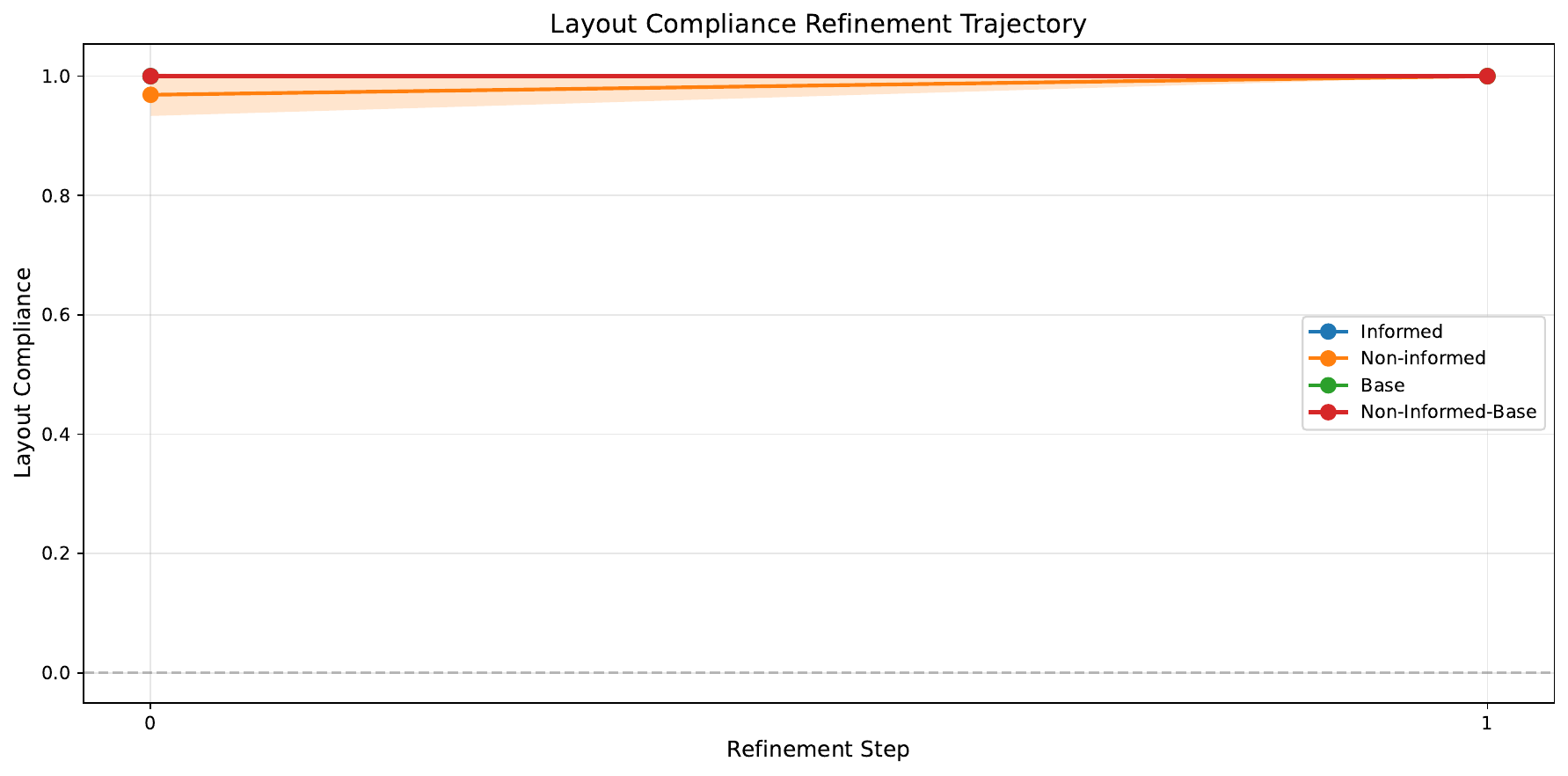}
        \caption{Layout Compliance}
    \end{subfigure}

    \vspace{1em}

    \begin{subfigure}{0.47\textwidth}
        \centering
        \includegraphics[width=\linewidth]{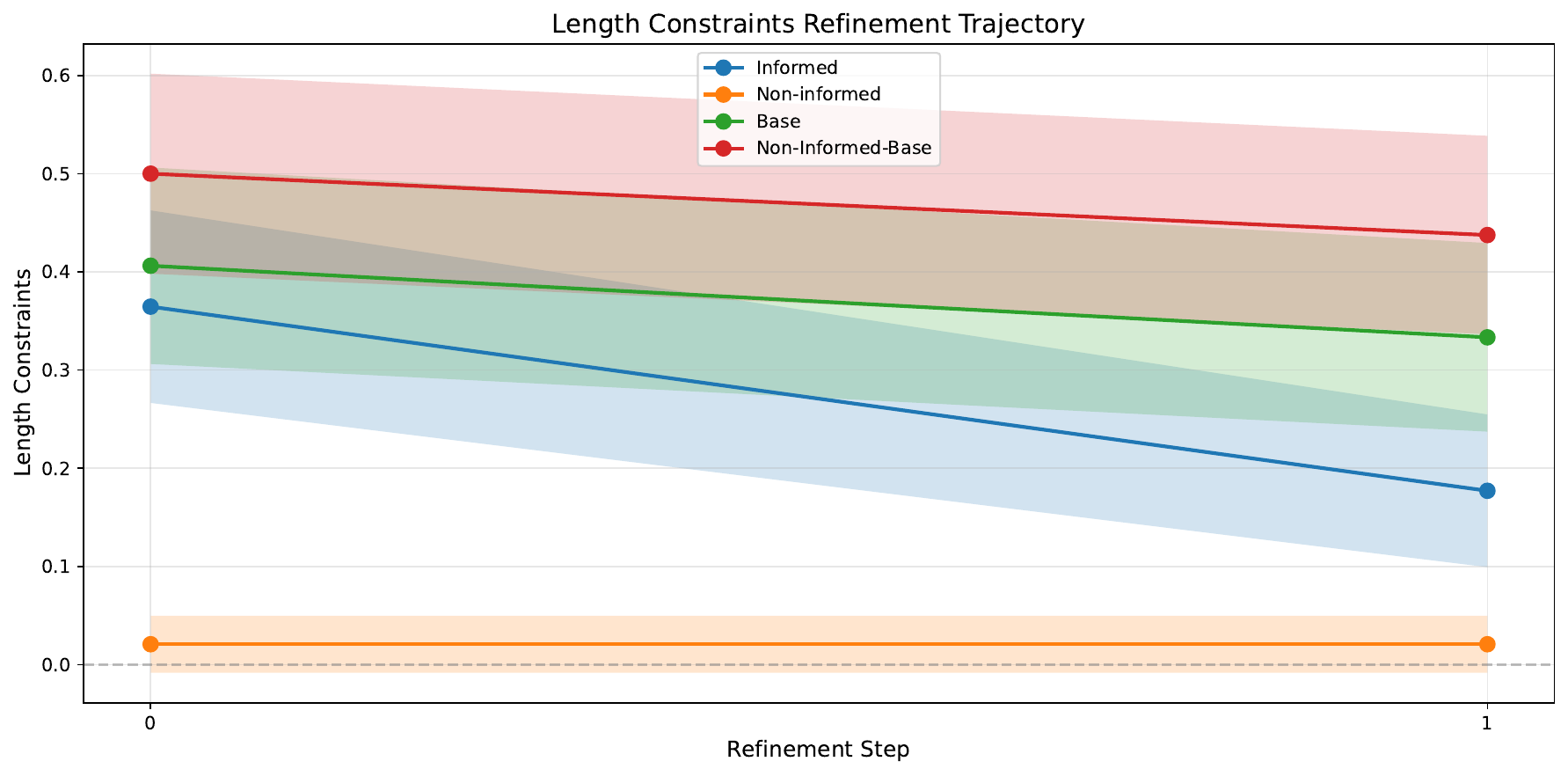}
        \caption{Length Constraints}
    \end{subfigure}%
    \hfill
    \begin{subfigure}{0.47\textwidth}
        \centering
        \includegraphics[width=\linewidth]{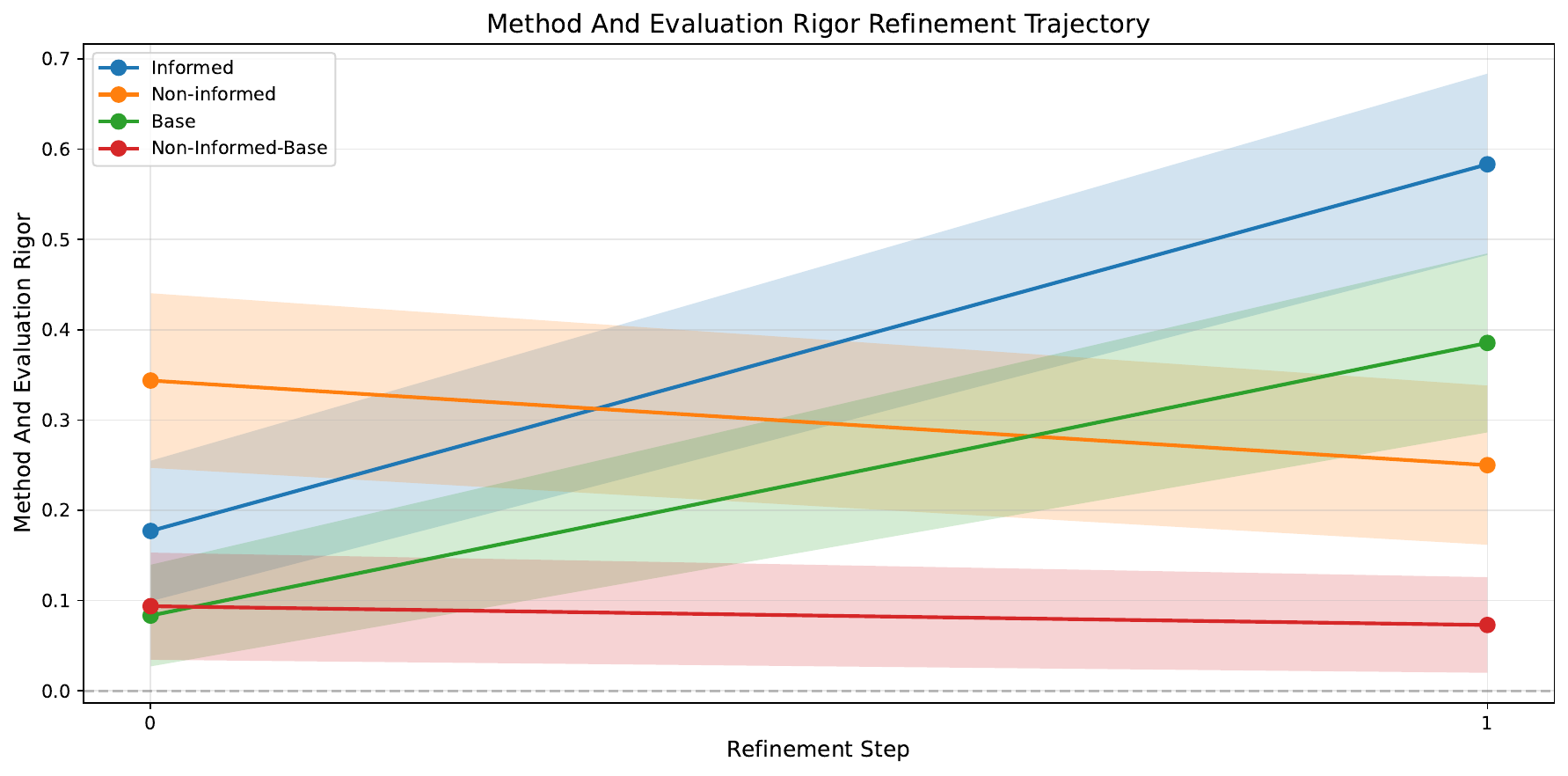}
        \caption{Method and Evaluation Rigor}
    \end{subfigure}

    \vspace{1em}

    \begin{subfigure}{0.47\textwidth}
        \centering
        \includegraphics[width=\linewidth]{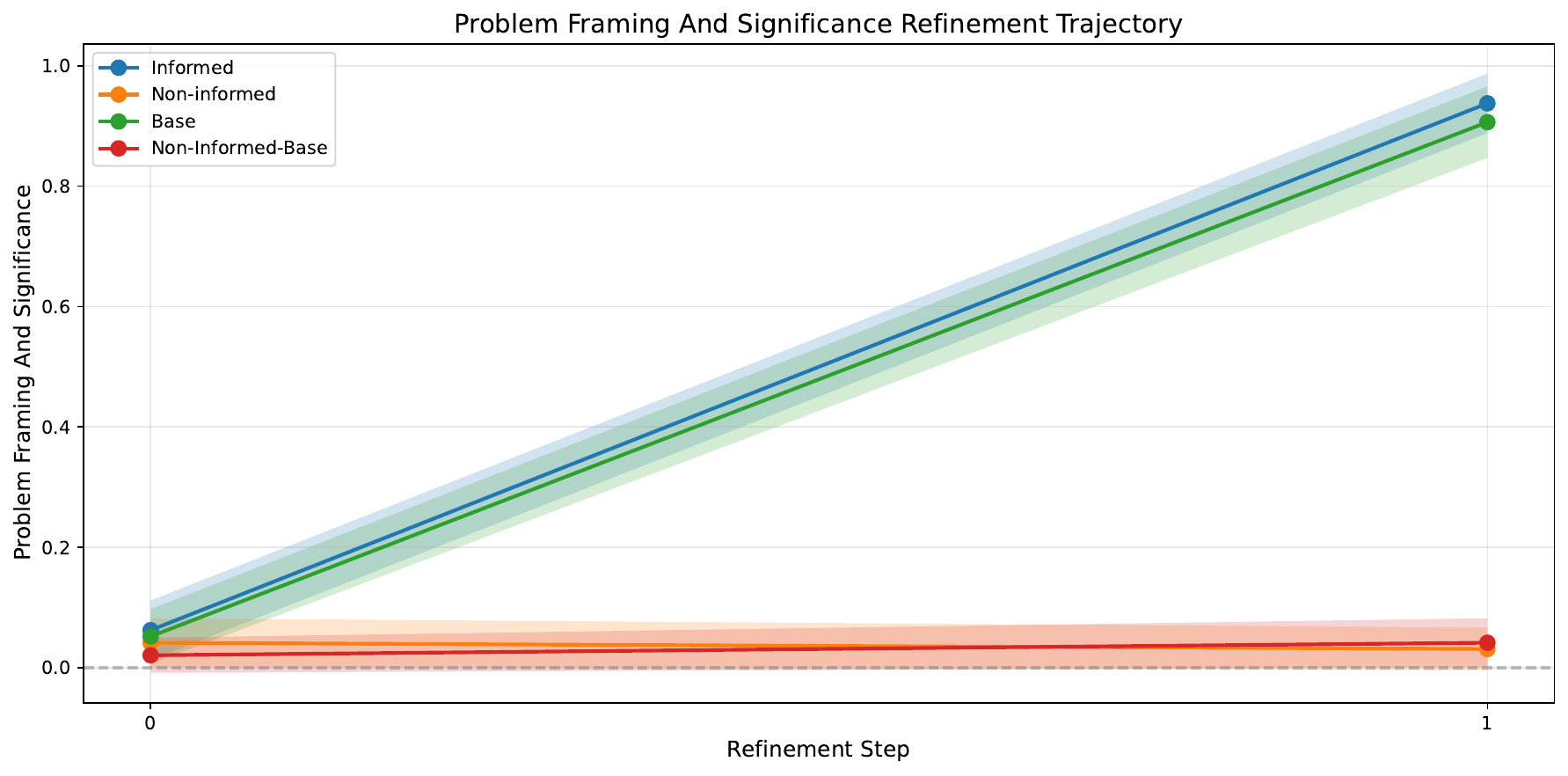}
        \caption{Problem Framing and Significance}
    \end{subfigure}%
    \hfill
    \begin{subfigure}{0.47\textwidth}
        \centering
        \includegraphics[width=\linewidth]{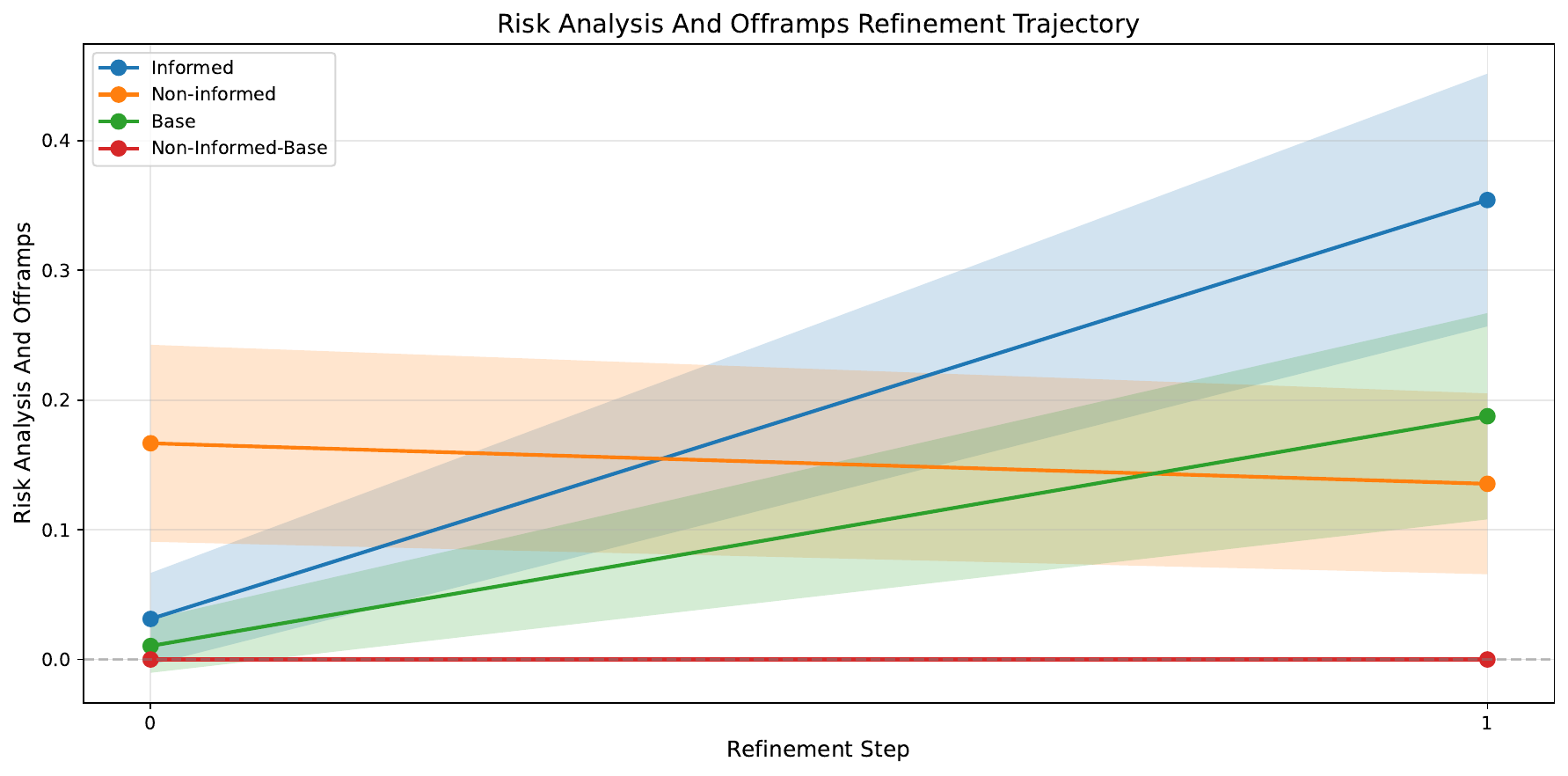}
        \caption{Risk Analysis and Offramps}
    \end{subfigure}

    \vspace{1em}

    \begin{subfigure}{0.47\textwidth}
        \centering
        \includegraphics[width=\linewidth]{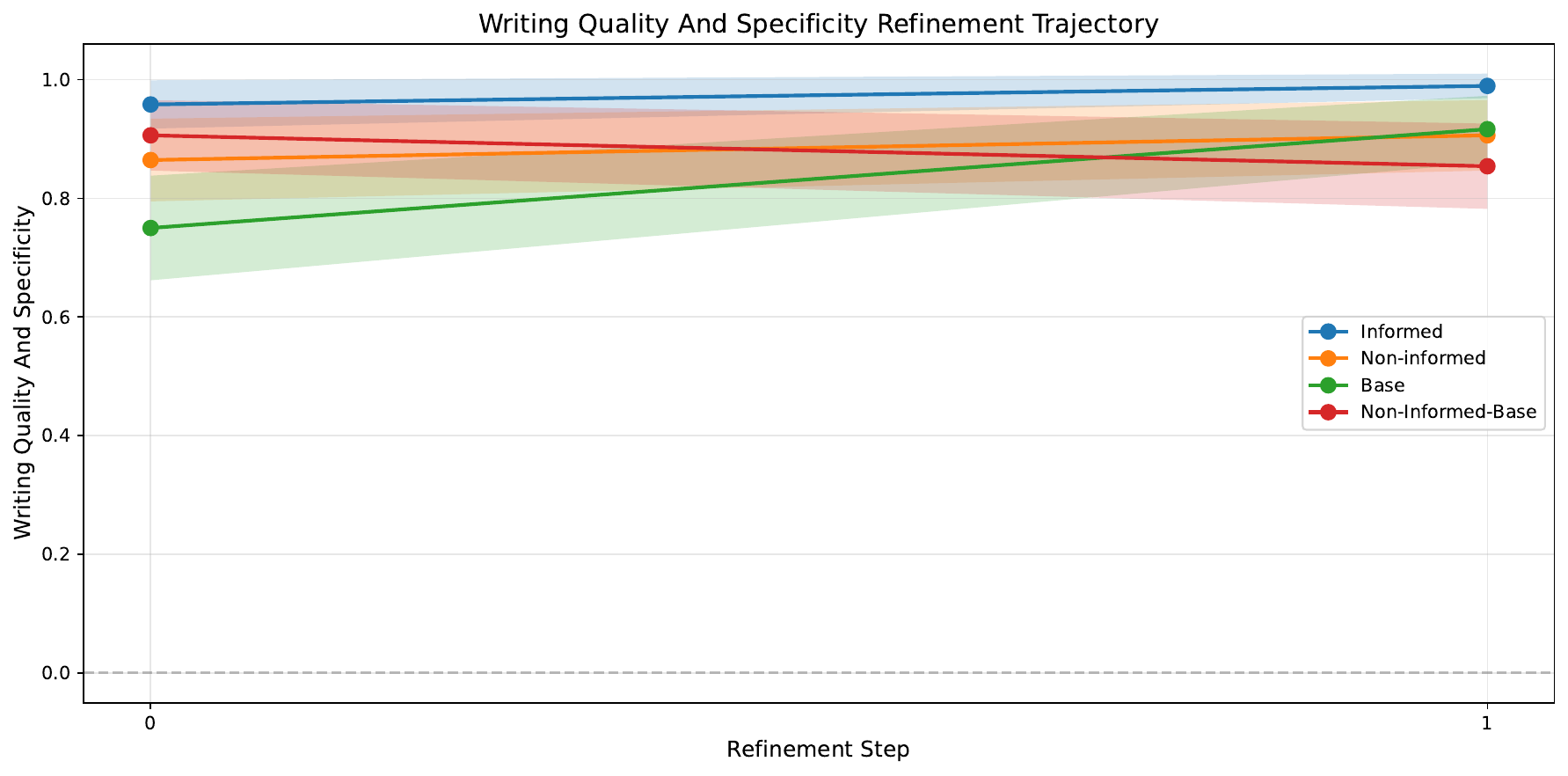}
        \caption{Writing Quality}
    \end{subfigure}

    \caption{Score trajectories for each checklist criteria. Informed (blue) is our model that has been trained with textual gradients and receives them during training; Non-Informed (orange) is our model that has been trained without textual gradients and is not receiving them during inference; Base (green) is the base model (Qwen3-4B-Thinking-2507) that receives textual gradients during inference; Non-Informed base is the base model that does not receive tetual gradients during inference. }
    \label{fig:appendix_synergies_grid}
\end{figure*}

Figure~\ref{fig:appendix_synergies_grid} shows the comparison described in Section~\ref{sec:additive_gains} for each checklist item separately. As for the aggregate figure in the main paper, we can observe a similar pattern per item, where the models that have access to textual gradients increase their score in the second step, while the non-informed ones do not. In addition, the same observed pattern of trained models outperforming their untrained counterparts also largely holds for the individual criteria, especially for the primary ones.

\end{document}